\documentclass[11pt]{article}

\usepackage[final]{acl}

\usepackage{times}
\usepackage{latexsym}
\usepackage{amssymb}
\usepackage[T1]{fontenc}

\usepackage[utf8]{inputenc}

\usepackage{microtype}

\usepackage{inconsolata}
\usepackage{amsmath} 
\usepackage{graphicx}
\usepackage{multirow}
\usepackage{booktabs}
\title{IRIS: Interleaved Reinforcement with Incremental Staged Curriculum for Cross-Lingual Mathematical Reasoning}

\author{
Navya Gupta\textsuperscript{1}\thanks{Equal contribution},
Rishitej Reddy Vyalla\textsuperscript{2}\footnotemark[1],
Avinash Anand\textsuperscript{1},
Chhavi Kirtani\textsuperscript{3},
Erik Cambria\textsuperscript{4},
\\
\textbf{Zhengchen Zhang}\textsuperscript{1},
\textbf{Zhengkui Wang}\textsuperscript{1},
\textbf{Timothy Liu}\textsuperscript{5},
\textbf{Aik Beng Ng}\textsuperscript{5},
\textbf{Simon See}\textsuperscript{5},
\textbf{Rajiv Ratn Shah}\textsuperscript{2}
\\
\\
\textsuperscript{1}Singapore Institute of Technology, Singapore \\
\textsuperscript{2}IIIT Delhi, New Delhi, India \\
\textsuperscript{3}University of California, San Diego, USA \\
\textsuperscript{4}Nanyang Technological University, Singapore \\
\textsuperscript{5}NVIDIA \\
}

\begin{document}
\maketitle
\begin{abstract}
Curriculum learning helps language models tackle complex reasoning by gradually increasing task difficulty. However, it often fails to generate consistent step-by-step reasoning, especially in multilingual and low-resource settings where cross-lingual transfer from English to Indian languages remains limited. We propose \textbf{IRIS: Interleaved Reinforcement with Incremental Staged Curriculum}, a two-axis framework that combines Supervised Fine-Tuning on progressively harder problems (vertical axis) with Reverse Curriculum Reinforcement Learning to reduce reliance on step-by-step guidance (horizontal axis). We design a composite reward combining correctness, step-wise alignment, continuity, and numeric incentives, optimized via Group Relative Policy Optimization (GRPO). We release \textbf{CL-Math}, a dataset of 29k problems with step-level annotations in English, Hindi, and Marathi. 
Across standard benchmarks and curated multilingual test sets, IRIS consistently improves performance, with strong results on math reasoning tasks and substantial gains in low-resource and bilingual settings, alongside modest improvements in high-resource languages. Our code and dataset will be publicly available at 
\url{https://github.com/avinanand/IRIS-Interleaved-Reinforcement-}.

\end{abstract}

\section{Introduction}
Mathematical reasoning challenges Large Language Models(LLMs) because correct final answers alone do not ensure valid reasoning. While recent proprietary models have made substantial progress in mathematical reasoning, improving smaller and open models, particularly in multilingual and low-resource settings, remains an open challenge. This gap highlights the limitations of training schemes that reward only end-answer accuracy, offering little supervision for the reasoning process itself.

Two complementary strands of work have emerged to address this limitation. At the step level, Reverse Curriculum Reinforcement Learning (R$^3$)~\citep{tao2024reverseforwardcurriculumlearning} emerges as an outcome supervision-based method where models complete partial reasoning chains and are rewarded based on final answer correctness. At the problem level,~\citet{anand2025multilingualmath} proposes a curriculum learning strategy that organizes math problems by difficulty, allowing models to progressively enhance reasoning capability in both English and bilingual settings. 

Problem-level curricula fail to capture where reasoning breaks down, while inference-time methods like Step-Guided Reasoning~\citep{zhang2023automatic} and Stepwise Self-Consistent CoT~\citep{zhu2023selfconsistentcot} show that supervising intermediate steps improves accuracy. This motivates integrating such fine-grained feedback directly into training. Conversely, R$^3$ typically trains on low-complexity datasets without difficulty scheduling, often leading to unstable learning under sparse rewards. Curriculum RL research~\citep{bengio2009curriculum, florensa2017reversecurriculum, parashar2024curriculumrl} shows that progressive task difficulty improves reward shaping and convergence.

\begin{figure*}[ht]
  \centering
  \includegraphics[width=\textwidth]{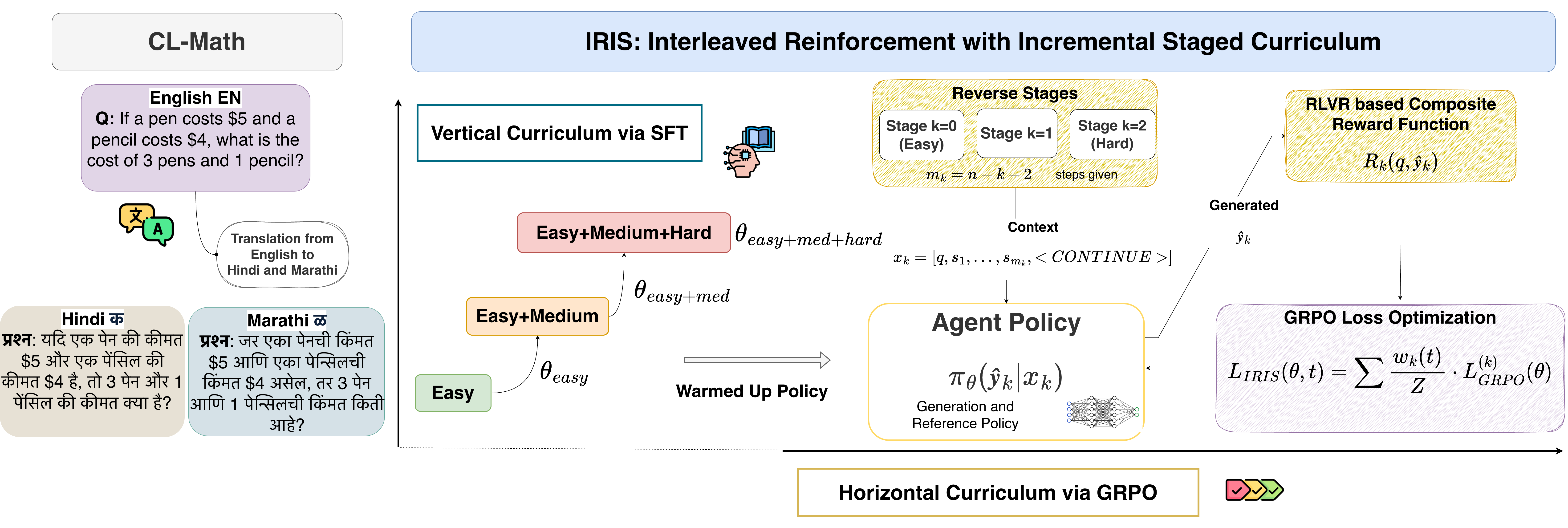} 
  \caption{ \textbf{Integrated IRIS: Interleaved Reinforcement with Incremental Staged Curriculum pipeline:} 
  It blends vertical supervised fine-tuning (SFT) with horizontal GRPO-based reinforcement learning. Starting with multilingual step-by-step data, the model is first warmed up on an easy-to-hard vertical curriculum, then refined through reverse-stage prompts and composite rewards to master deep, robust mathematical reasoning.
  }
  \label{fig:basic_pipeline}
\end{figure*}

To address this challenge, we propose \textbf{IRIS: Interleaved Reinforcement with Incremental Staged Curriculum}, targeting performance enhancements in small language models (SLMs) (see Figure \ref{fig:basic_pipeline} for detailed architecture), which pairs a problem-level curriculum with step-level feedback, using RL to guide the model from supervised steps to fully independent, multi-step solutions. This two-axis curriculum mirrors human cognitive development~\citep{bengio2009curriculum}, first to understand reasoning structure, then scaling to harder and longer tasks.

To support learning under this structured curriculum, we introduce a composite, rule-checked reward signal, that supervises both intermediate steps and final answers within our curriculum. Building on prior work that has provided supervision at individual granularities, this multi-part reward structure provides richer supervision and a more stable learning signal.
For RL optimization, we adopt Group Relative Policy Optimization (GRPO)~\citep{shao2024deepseekmath}, designed for training Deepseek's R1 reasoning models. It has been shown to outperform PPO/DPO with increased efficiency with no reward or value model involved under compositional reward settings.

Finally, to evaluate generalization across languages, we release \textbf{CL-Math}, a novel 29k-example multilingual dataset with step-level annotations in English, Hindi, and Marathi, enabling bilingual and low-resource training for underrepresented languages. It consists of 2184 easy, 2212 medium, and 5588 hard samples in all three languages. IRIS extends beyond English by incorporating both monolingual (Hindi, Marathi) and bilingual (English–Hindi, English–Marathi) setups, enabling controlled analysis of multilingual reasoning and cross-lingual transfer. Bilingual training allows shared representations from English to reinforce reasoning quality in low-resource languages, while monolingual setups isolate native-language learning dynamics.

We benchmark our method on CL-Math and its Hindi and Marathi variants. Compared to strong baselines, IRIS improves reasoning accuracy by up to +15.8\% on the hardest Hindi problems as well as on the English–Hindi Medium+Hard bilingual evaluation, clearly demonstrating that our two-axis curriculum learning yields more robust reasoning across languages and problem complexities. Beyond in-domain gains, \textbf{IRIS (Qwen2.5-Math-7B)} also achieves 90.6\% on SVAMP~\citep{patel-etal-2021-nlp}, 83.9\% on GSM8K~\citep{cobbe2021gsm8k}, and 64.6\% on MATH~\citep{hendrycks2021measuringmathematicalproblemsolving}, surpassing the base model results and other specialized 7B models.

This work makes the following key contributions:

\begin{itemize}
  \item We introduce a \textbf{IRIS: Interleaved Reinforcement with Incremental Staged Curriculum} framework that jointly leverages difficulty-based progression across problems and partial-solution supervision within problems to improve mathematical reasoning in SLMs.
  
  \item We design a \textbf{composite reward structure} that integrates multiple feedback signals offering rich supervision during RL optimization.

  \item We construct \textbf{CL-Math}, a new \textbf{multilingual math reasoning and curriculum learning dataset} comprising 29k examples with fine-grained reasoning annotations in English, Hindi, and Marathi, supporting both monolingual and bilingual training setups.
  
  \item We conduct extensive experiments showing that our approach consistently outperforms baselines on both standard and multilingual benchmarks, demonstrating enhanced depth of reasoning and improved cross-lingual generalization.

\end{itemize}

\section{Related Work}
Mathematical reasoning represents a challenging intersection of natural language understanding and computational mathematics for LLMs~\citep{wang2024mathcoder,nlu,ying2024internlm,yu2023metamath} and open-source SLLMs~\citep{guan2025rstar,kai2024logic}. While early gains were largely scale-driven, recent work emphasizes structured, step-by-step reasoning using methods like Chain-of-Thought prompting and self-consistency~\citep{wei2023chainofthoughtpromptingelicitsreasoning,wang2023selfconsistencyimproveschainthought}. Curriculum learning and reinforcement learning have emerged as two powerful paradigms to further enhance this structured reasoning ability, by scheduling data along a progressive difficulty gradient and enabling models to explore reasoning trajectories through interaction-based feedback~\citep{zhagra,tuuada}. Multilingual extension of these methods to low-resource Indian languages is also critical, where limited data and linguistic variability pose unique challenges~\citep{anand2025multilingualmath, dnyanesh_walwadkar_2024, sharma-etal-2022-hawp}. 


\subsection{Curriculum Learning with Structured Solutions}
Curriculum learning, first formalized by~\citet{bengio2009curriculum}, has shown consistent benefits for complex tasks by training models on increasingly difficult problems~\citep{pattnaik-etal-2024-enhancing,Moein_2024,soviany2022curriculum}. In the context of mathematical reasoning, organizing questions by difficulty and fine-tuning with structured solutions improves both accuracy and stability~\citep{anand2025multilingualmath}. However, extending this approach to multilingual settings is challenging due to data imbalance and cross-linguistic variation, which can degrade curriculum coherence and reward quality.



\subsection{Reinforcement Learning for Mathematical Reasoning}
Reinforcement learning (RL) enables models to explore and self-correct beyond supervised imitation~\citep{bai2022,ouyang2022traininglanguagemodels,havrilla2024teaching}. Step-level methods such as R$^3$~\citep{tao2024reverseforwardcurriculumlearning} and verifier-based frameworks like RLVR~\citep{lightman2023letsverifystepstep,zhang2025generativeverifiersrewardmodeling,xiong2025selfrewardingcorrectionmathematicalreasoning} provide richer intermediate feedback. However, most prior work applies these ideas independently, lacking structured progression across task difficulty. Our work unifies both by combining interleaved curricula with verifier-informed rewards optimized via GRPO~\citep{shao2024deepseekmath} for more stable credit assignment.
\subsection{Multilingual Learning and Cross-Lingual Transfer in LLMs}

Extending reasoning to multilingual contexts, especially underrepresented languages, remains an active area of research. Recent studies focus on creating bilingual datasets~\citep{anand2025multilingualmath}, developing evaluation benchmarks~\citep{marchisio-etal-2024-understanding,zhang2023m3exammultilingualmultimodalmultilevel,iyer2025xl}, and leveraging cross-lingual transfer~\citep{shaham2024multilingual} or translation-based augmentation~\citep{wang-etal-2023-self-augmentation}. However, persistent challenges remain in achieving equitable multilingual prowess for low-resource Indian languages. Our work addresses this gap by integrating bilingual curricula and evaluating transfer across English, Hindi, and Marathi.

\section{Methodology}

\subsection{Dataset Curation}
We introduce \textbf{CL-Math}, a new multilingual corpus that extends the benchmark proposed by \citet{anand2025multilingualmath}. They unify existing benchmarks to form \textit{IndiMathQA} having three difficulty tiers(\emph{Easy}, \emph{Medium}, \emph{Hard}) to support curriculum-based fine-tuning. \textbf{CL-Math} extends this framework to a multilingual setting and adds structured reasoning. To ensure annotation quality, the dataset was independently validated using Fleiss’ kappa, confirming high inter-annotator agreement. Evaluation-time agreement is reported separately in Appendix.

\subsubsection{Difference from IndiMathQA}
IndiMathQA consists of question–answer pairs without intermediate reasoning annotations. In contrast, CL-Math augments each problem with fine-grained, step-wise reasoning traces generated using a carefully prompted Llama-3.3-70B model~\citep{grattafiori2024llama} conditioned on human-verified final answers, guided by few-shot examples and carefully designed prompts tailored to the problem complexity. These steps are structured over validated explanations rather than free-form generations.


\subsubsection{Multilingual Translation}
Next, problem statements and their structured solutions were automatically translated into Hindi and Marathi using AI4Bharat’s \textit{IndicTrans2} (Indic-En,1.1B)~\citep{gala2023indictrans}, ensuring that each reasoning step remains intact across languages, validated by human verification. Importantly, we preserved the train/test split prior to translation, and each language version was generated separately from these fixed splits. This ensured that no translated instance from the English training set could appear in the test sets of Hindi or Marathi, eliminating any potential data leakage across languages.
By combining granular solution paths with high-quality translations, \textit{CL-Math} supports interleaved curriculum learning in both English and two major Indian languages. This design enables models to progress through mathematical reasoning tasks in a linguistically diverse environment.

\subsection{IRIS: Interleaved Reinforcement with Incremental Staged Curriculum}

In this section, we present the IRIS: Interleaved Reinforcement with Incremental Staged Curriculum, which structures training along two complementary axes. The Vertical Axis stages supervised fine‑tuning from easy to hard problems, while the Horizontal Axis applies step‑wise continuation via reinforcement learning to push reasoning beyond given solution prefixes. We first describe the vertical curriculum, then the horizontal continuation setup, and finally show how these two stages interleave. We conclude by demonstrating how the same pipeline extends to Hindi and Marathi for multilingual mathematical reasoning.

\subsection{Vertical Axis: Problem-Wise Curriculum Learning}
\label{subsec:vertical-axis}

The vertical axis applies a difficulty-based curriculum over \textbf{CL-Math} ($\mathcal{D}$), using supervised fine-tuning to expose the model first to short, low-complexity reasoning traces before progressing to longer solutions.

Starting from the pretrained model parameters $\theta^{(0)}$, we perform sequential SFT over increasing difficulty levels. Training on $\mathcal{D}_{\text{easy}}$ yields $\theta_{\text{easy}}$, which is then further fine-tuned on $\mathcal{D}_{\text{medium}}$ to obtain $\theta_{\text{easy+med}}$, and finally on $\mathcal{D}_{\text{hard}}$ to produce $\theta_{\text{easy+med+hard}}$. At each stage, supervision is applied over full serialized ground-truth reasoning traces using standard cross-entropy loss.

We denote the resulting checkpoint at curriculum stage $c$ as $\theta_c$, where
\(
c \in \{\text{easy},\,\text{easy+med},\,\text{easy+med+hard}\}.
\)
These checkpoints serve as initialization for the Horizontal Axis GRPO-based reverse curriculum described next.

\subsection{Horizontal Axis : Step-Wise Continuation Curriculum}

\subsubsection{Motivation for RL}
Supervised fine‑tuning provides a strong starting point by teaching the model to follow reasoning patterns observed in the training data. Reinforcement learning builds on this foundation, rewarding correct and novel continuations, and enabling the model to extend reasoning beyond the typical chains seen during SFT.

\subsection{Continuation Setup}

We establish the following notation for the horizontal curriculum:

\begin{itemize}
    \item \textbf{Question}: $q$ denotes the problem statement
    \item \textbf{Ground-truth solution}: $\mathbf{y} = [s_1, s_2, \ldots, s_n]$ where $s_i$ represents the $i$-th reasoning step, and $s_n$ includes the final answer
    \item \textbf{Curriculum stage}: $k \in \{0, 1, \ldots, n-2\}$ indexes the curriculum difficulty, where higher $k$ corresponds to harder tasks
    \item \textbf{Prefix length}: At stage $k$, we provide $m_k = n - k - 2$ ground-truth steps as context. Henceforth, the input prompt becomes $\mathbf{x}_k = [q, s_1, \ldots, s_{m_k}, \texttt{<CONTINUE>}]$ representing the complete context given to the model, consisting of the question, the first $m_k$ reasoning steps, and the special continuation token
    \item \textbf{Generated suffix}: $\hat{\mathbf{y}}_k = [\hat{s}_{m_k+1}, \ldots, \hat{s}_n]$ denotes the model's generated continuation
\end{itemize}

The \texttt{<CONTINUE>} token serves as an explicit marker that signals the model to generate the remaining solution steps beyond the provided prefix. This token separates the given context from what the model must produce, enabling clear delineation between the provided steps and model-generated reasoning. At stage $k=0$ (easiest), the model receives the question plus $n-2$ steps and must generate only the final 2 steps. At stage $k=n-2$ (hardest), only the question is provided ($m_k = 0$), and the model must produce the complete $n$-step solution. 

\begin{figure}[htbp]
  \centering
  \includegraphics[width=\linewidth]{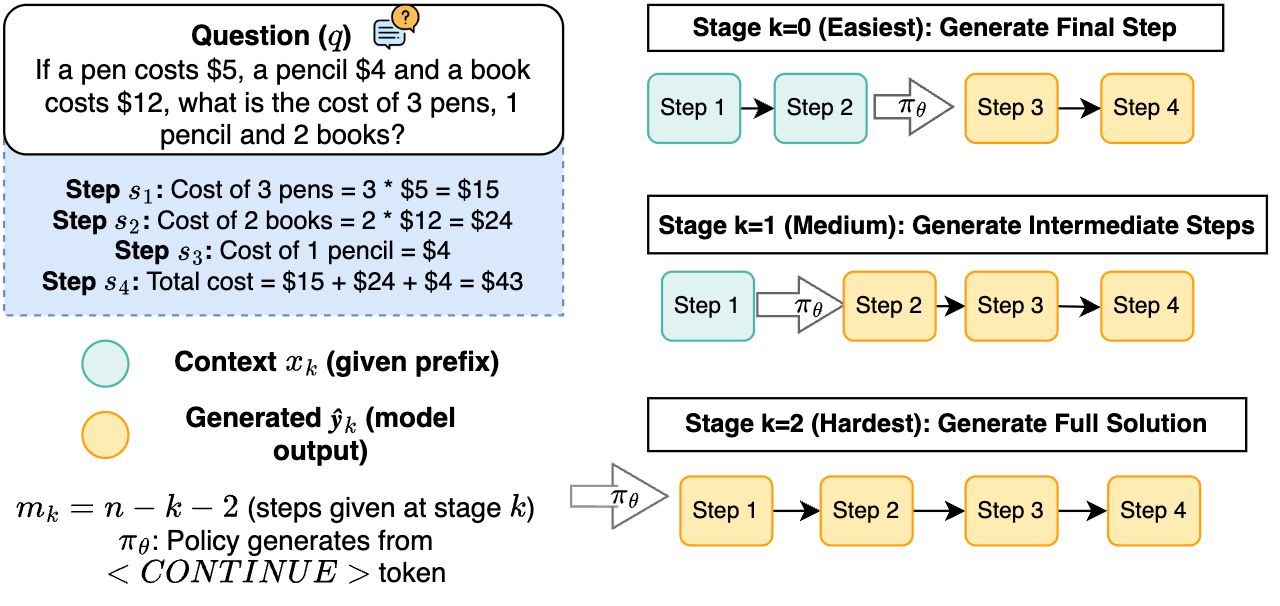} 
  \caption{\textbf{Reverse curriculum staging.} Progressive removal of reasoning context guides the model from partial completion to full solution generation.}
  \label{fig:stages}
\end{figure}

\textbf{Reverse Curriculum Staging.} Each problem $(q, \mathbf{y})$ with $n$ reasoning steps is decomposed into $n-1$ training instances, one per stage $k \in \{0, \ldots, n-2\}$. Figure~\ref{fig:stages} illustrates this decomposition: as $k$ increases, the prefix shrinks and the model must generate longer continuations, creating a reverse curriculum from easy (finishing nearly-complete solutions) to hard (generating full solutions from scratch).

We deliberately define the stage range as $\{0, \ldots, n-2\}$ rather than $\{0, \ldots, n-3\}$ to ensure that even problems with very short solution chains yield at least one nontrivial training instance. We exclude $k = n-1$ because that would provide all steps but the final answer, leaving nothing for the model to generate.

Formally, the model learns the conditional distribution:
\begin{equation}
    \pi_\theta(\hat{\mathbf{y}}_k \mid \mathbf{x}_k) = \pi_\theta(s_{m_k+1}, \ldots, s_n \mid q, s_1, \ldots, s_{m_k})
\end{equation}

This curriculum schedule allows the model to first master short-range reasoning before progressively learning to plan deeper from sparser context.


\subsection{RLVR-Based Composite Reward Design}

In the horizontal axis, we define a composite reward function that promotes answer correctness, alignment with reference reasoning, solution continuity, and well-formed numeric outputs. For a given problem $(q, \mathbf{y})$ at stage $k$, each model rollout $\hat{\mathbf{y}}_k$ receives:

\begin{equation}
R_k(q, \hat{\mathbf{y}}_k) = r_{\text{final}} + \lambda_k \cdot r_{\text{cos}} + r_{\text{cont}} + r_{\text{int}}
\end{equation}

where each component is defined as follows:

\textbf{Correctness Reward.} 
\(r_{\text{final}}\in\{+2,0\}\) assigns +2 for an exact answer match and 0 otherwise. 
This binary reward provides the primary learning signal, assigning full credit only when the generated final answer matches the ground truth exactly.

\textbf{Cosine Alignment Reward.}
\(r_{\text{cos}} \in [0,1]\)

This term measures semantic similarity between the generated suffix and the ground-truth continuation steps from CL-Math. These continuation steps are structured reformattings of pre-validated explanations and are used solely as a stabilization signal during early training. Importantly, this reward does not distill external model reasoning or impose teacher-generated solutions; instead, it encourages faithful continuation of already verified solution trajectories, reducing variance in early GRPO updates.

It is computed using SentenceTransformers encoders: \textit{all-MiniLM-L6-v2} for English and \textit{LaBSE} for Hindi and Marathi, ensuring consistent multilingual semantic alignment.

\textbf{Stage-Dependent Alignment Weight.}
\begin{equation}
\lambda_k = \lambda_{\max} \left(1 - \frac{k}{k_{\max}}\right), \quad k_{\max} = n-2
\end{equation}

Here, $\lambda_{\max}$ is the maximum alignment weight (set to 2.5), and $\lambda_k$ decays linearly as stage $k$ increases. At stage $k=0$, when reasoning traces are only lightly truncated, cosine alignment provides a stabilizing scaffold with full weight $\lambda_0 = \lambda_{\max}$. As the curriculum progresses to higher stages where fewer steps are provided, $\lambda_k$ decreases, reducing the influence of alignment. At the hardest stage $k = k_{\max}$, we have $\lambda_{k_{\max}} = 0$, eliminating alignment entirely.

This schedule ensures that cosine alignment serves as a transient aid in early stages, guiding the model toward coherent continuation behavior. As the model advances through the curriculum, the dominant signal shifts to correctness ($r_{\text{final}}$), encouraging the model to discover valid final answers even if its reasoning path differs from the reference. 
\textbf{Continuation Reward.} $r_{\text{cont}}\!\in\{-0.5, 0, +1\}$

This reward encourages proper step numbering: the model should continue from where the prefix ended rather than restarting the step count. This promotes structural coherence in multi-step reasoning.

\textbf{Integer Reward.} \(r_{\text{int}}\!\in\{+0.5,0\}\)

This provides a small bonus whenever the final answer token represents any integer, encouraging the policy to output numeric results even before learning to match exact values.

Both $r_{\text{cont}}$ and $r_{\text{int}}$ are deliberately small compared to $r_{\text{final}}$ to shape behavior without distorting the primary learning signal. Together, these four components balance immediate correctness with structural reasoning quality, while the stage-dependent weight $\lambda_k$ ensures that exploration dominates in later curriculum stages.

\subsection{Group Relative Policy Optimization}

We optimize the policy $\pi_\theta$ using Group Relative Policy Optimization (GRPO) \citep{shao2024deepseekmath}. For each problem $(q, \mathbf{y})$ at stage $k$, we sample $G$ rollouts from the current policy and compute advantages relative to the group mean. The GRPO objective incorporates these advantages along with a KL penalty to maintain stability. For the complete loss formulation, refer to Appendix.

\textbf{Curriculum-Weighted Sampling.} To align stage sampling with the model's evolving capabilities, we assign each stage $k$ a time-varying weight:

\[
w_k(t) = \alpha_t^k,\quad  \alpha_t = \alpha_0 + (\alpha_1 - \alpha_0) \cdot \min\left(\frac{t}{T}, 1\right)
\]

Here, $t$ is the current training step, $T$ is the warm-up period (10\% of total updates), $\alpha_0 = 0.7$, and $\alpha_1 = 1.0$. At $t=0$, this exponentially favors easier stages (small $k$); by $t=T$, sampling becomes uniform across all stages.

The overall IRIS loss integrates these curriculum weights:

\begin{equation}
\mathcal{L}_{\text{IRIS}}(\theta, t) = \sum_{k=0}^{n-2} \frac{w_k(t)}{\sum_{j=0}^{n-2} w_j(t)} \cdot \mathcal{L}_{\text{GRPO}}^{(k)}(\theta)
\end{equation}

This weighted combination trains on a time-varying mixture of stage difficulties, integrating our stage-dependent reward design with GRPO optimization.







\subsection{Extending IRIS to the Multilingual Setting}
\label{sec:multilingual}

Language models that \emph{reason} should do so regardless of the script in which the question is asked. 
To verify that our two-axis curriculum scales beyond English, we replicate the entire pipeline for
\textbf{Hindi}$\sim$HI and \textbf{Marathi}$\sim$MR and then explore bilingual transfer.

\paragraph{Monolingual setting.}
For each language~$\ell\!\in\!\{\text{HI},\text{MR}\}$ we create difficulty partitions
$\mathcal{D}_{\text{easy}}^{(\ell)},\,
 \mathcal{D}_{\text{medium}}^{(\ell)},\,
 \mathcal{D}_{\text{hard}}^{(\ell)}$
by direct translation from their English counterparts.

We then apply \textbf{IRIS} in the same manner,
the result is a trio of language-specific models
\(\{\theta_{\text{easy}}^{(\ell)},\,
  \theta_{\text{easy+med}}^{(\ell)},\,
  \theta_{\text{easy+med+hard}}^{(\ell)}\}\),
each further refined by horizontal GRPO.

\textbf{Bilingual Cross-Transfer.} We test whether mixing languages accelerates learning by creating balanced bilingual datasets:
\(\mathcal{D}^{\text{EN+HI}}, \quad \mathcal{D}^{\text{EN+MR}}\)
each stratified into easy, medium, and hard partitions. We apply the IRIS framework to both bilingual splits and evaluate cross-lingual transfer capabilities; the empirical effect of this design choice is analysed in Section~\ref{sec:anchor}.

\subsection{Core Design Principles}
The following points \textbf{summarize} the key design principles underlying the proposed training framework, which are empirically validated in Section~\ref{sec:ablation}:

\begin{itemize}
    \item \textbf{IRIS} is \textbf{centered around curriculum-driven reinforcement learning}, where both task difficulty and reasoning horizon are \textbf{progressively controlled over time}, enabling stable acquisition of long-chain reasoning without direct training on full-length hard traces.

    \item A problem-wise curriculum is used to \textbf{explicitly control when the model is exposed to difficult data}, rather than relying on mixed or hard-only training distributions.

    \item The composite reward is \textbf{designed to shape intermediate reasoning behavior}, reflecting the view that final-answer-only rewards are insufficient for optimizing complex multi-step solutions.

    \item These training principles are \textbf{applied consistently across languages}, enabling evaluation of their robustness beyond English-centric settings.
\end{itemize}

\section{Experimental Setup} 
\label{sec:experiment-setup}

\subsection{Models and Implementation Details}

Our experiments primarily use \textbf{Qwen2.5-Math-7B}~\citep{yang2024qwen2}, evaluated under two settings: (1) curriculum-guided SFT and (2) stepwise curriculum reinforcement learning (SCRL). To assess generalizability, we also fine-tune \textbf{WizardMath-7B}~\citep{luo2023wizardmath} using the same framework.
In the first phase, we train for three epochs with a learning rate of $3\times10^{-4}$ using the \textit{AdamW} optimizer, a 10\% warm-up ratio, and gradient accumulation over 16 steps to ensure stable convergence. Training for the medium- and hard-level curricula resumes from the previous checkpoint to maintain curriculum continuity.
In the \textbf{Full IRIS (V+H)} phase, we apply Group Relative Policy Optimization (GRPO) with a reduced learning rate of $5\times10^{-6}$ and reward-weighted updates. Each prompt generates four completions per rollout for one epoch of reinforcement optimization. Cosine scheduling, 8-bit optimization, and vLLM-backed inference are used for efficient rollout and memory management; hardware details are provided in the Appendix.

\begin{table*}[t]
\centering
\small
\caption{ Zero-shot Pass@1 Accuracy (\%) for vertical-only (V) and vertical+horizontal (V+H) curricula. Models are evaluated on plain-language and bilingual benchmarks using 80-20 train-test split of CL-Math. Bilingual results show cross-lingual transfer to Hindi (HI) and Marathi (MR).}
\begin{tabular}{ll@{\hspace{4pt}}cc@{\hspace{4pt}}cc@{\hspace{4pt}}cc@{\hspace{4pt}}cc@{\hspace{4pt}}cc}
\toprule
\multirow{2}{*}{\textbf{Model}} & \multirow{2}{*}{\textbf{Level}} & 
\multicolumn{2}{c}{\textbf{EN}} & \multicolumn{2}{c}{\textbf{HI}} & \multicolumn{2}{c}{\textbf{MR}} & 
\multicolumn{2}{c}{\textbf{EN+HI}} & \multicolumn{2}{c}{\textbf{EN+MR}} \\
\cmidrule(lr){3-4} \cmidrule(lr){5-6} \cmidrule(lr){7-8} \cmidrule(lr){9-10} \cmidrule(lr){11-12}
& & V & V+H & V & V+H & V & V+H & V & V+H & V & V+H \\
\midrule
\addlinespace[3pt]
\multirow{3}{*}{\shortstack{\textbf{Qwen2.5-Math 7B}}} 
& Easy         & 85.2 & \textbf{87.0} & 52.7 & \textbf{65.5} & 38.3 & \textbf{41.0} & 56.4 & \textbf{69.0} & 51.6 & \textbf{57.5} \\
& Easy+Med       & 85.8 & \textbf{88.4} & 54.2 & \textbf{70.1} & 54.0 & \textbf{57.5} & 59.6 & \textbf{70.3} & 59.0 & \textbf{59.3} \\
& Easy+Med+Hard     & \textbf{86.3} & 85.3 & 57.7 & \textbf{73.5} & 55.7 & \textbf{59.0} & 61.2 & \textbf{77.0} & 61.0 & \textbf{64.2} \\
\addlinespace[3pt]
\multirow{3}{*}{\shortstack{\textbf{WizardMath-7B}}} 
& Easy         & 49.7 & \textbf{52.4} & 21.2 & \textbf{26.6} & 10.5 & \textbf{13.1} & 30.0 & \textbf{36.2} & 20.0 & \textbf{21.2} \\
& Easy+Med       & 52.0 & \textbf{53.4} & 25.0 & \textbf{35.3} & 14.0 & \textbf{27.2} & 32.8 & \textbf{37.2} & 27.0 & \textbf{30.0} \\
& Easy+Med+Hard     & 54.0 & \textbf{55.0} & 32.8 & 32.8 & 18.6 & \textbf{20.6} & 32.0 & \textbf{32.5} & 23.0 & \textbf{24.0} \\
\bottomrule
\end{tabular}
\label{tab:final_results}
\end{table*}

\begin{table*}[t]
\centering
\small
\caption{ Zero-shot Pass@1 Accuracy (\%) on SVAMP, GSM8K, and MATH benchmarks.}
\setlength{\tabcolsep}{4pt} 
\begin{tabular}{l@{\hspace{4pt}}c@{\hspace{4pt}}c@{\hspace{4pt}}c@{\hspace{4pt}}c@{\hspace{4pt}}c@{\hspace{4pt}}c@{\hspace{4pt}}c}
\toprule
\textbf{Benchmark} & 
\textbf{Qwen2.5 IRIS} & 
\textbf{WizMath IRIS} & 
\textbf{GPT-4} & 
\textbf{Qwen2.5 Base} & 
\textbf{WizMath Base} & 
\textbf{MetaMath} & 
\textbf{DeepSeek-Math} \\
\midrule
SVAMP   & \textbf{90.6} & \textbf{70.4} & 93.1 & 82.1 & 57.3 & 68.8 & 73.2 \\
GSM8K   & 83.9 & 70.6 & 92 & 79.3 & 54.9 & 66.5 & 64.2 \\
MATH    & \textbf{64.6} & \textbf{29.5} & 80 & 55.4 & 10.7 & 19.8 & 36.2 \\
\bottomrule
\end{tabular}
\label{tab:benchmark_results}
\end{table*}

\section{English as a Reasoning Anchor for Multilingual Transfer}
\label{sec:anchor}
\begin{figure}[htbp]
  \centering
  \includegraphics[width=\linewidth]{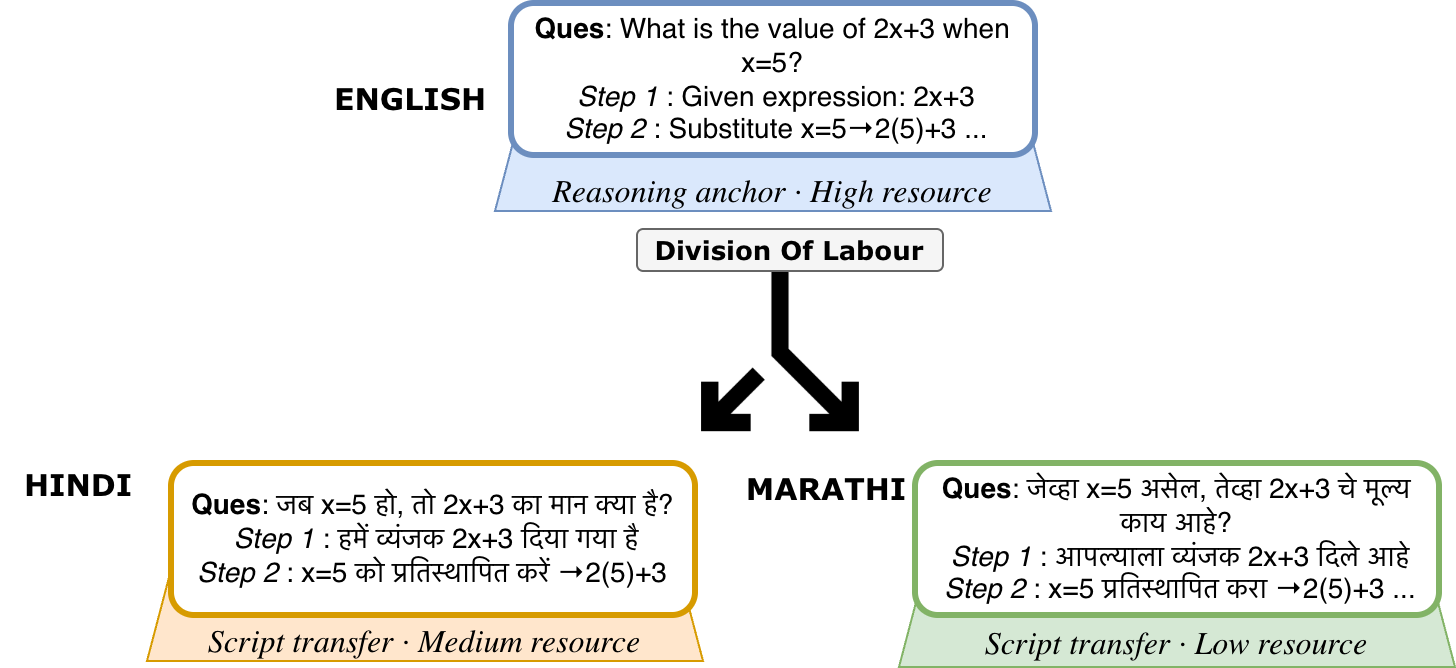} 
  \caption{\textbf{Training mix illustrating the division of labour:} English reinforces existing reasoning ability (high resource), while Hindi and Marathi inherit that reasoning and focus solely on expressing it in a new linguistic form (medium and low resource respectively).}
  \label{fig:Language Choice}
\end{figure}
Grounded in established principles of cross-lingual transfer learning and continual learning, we hypothesize that English acts as a reasoning anchor during multilingual RL post-training. While pivot language approaches and catastrophic forgetting mitigation are well-studied in static representation learning, their role in stabilizing RL-based mathematical reasoning transfer to low-resource Indic languages remains unexplored. As illustrated in Figure~\ref{fig:Language Choice}, this creates a natural division of labour: English reinforces the model's existing step-by-step reasoning ability, while Hindi and Marathi, sharing Devanagari script and related syntax, focus exclusively on expressing that reasoning in a new linguistic form. The model decouples the two problems entirely, inheriting its reasoning competence from English and transferring it across scripts with minimal additional burden.

\begin{figure}
    \centering
    \includegraphics[width=\linewidth]{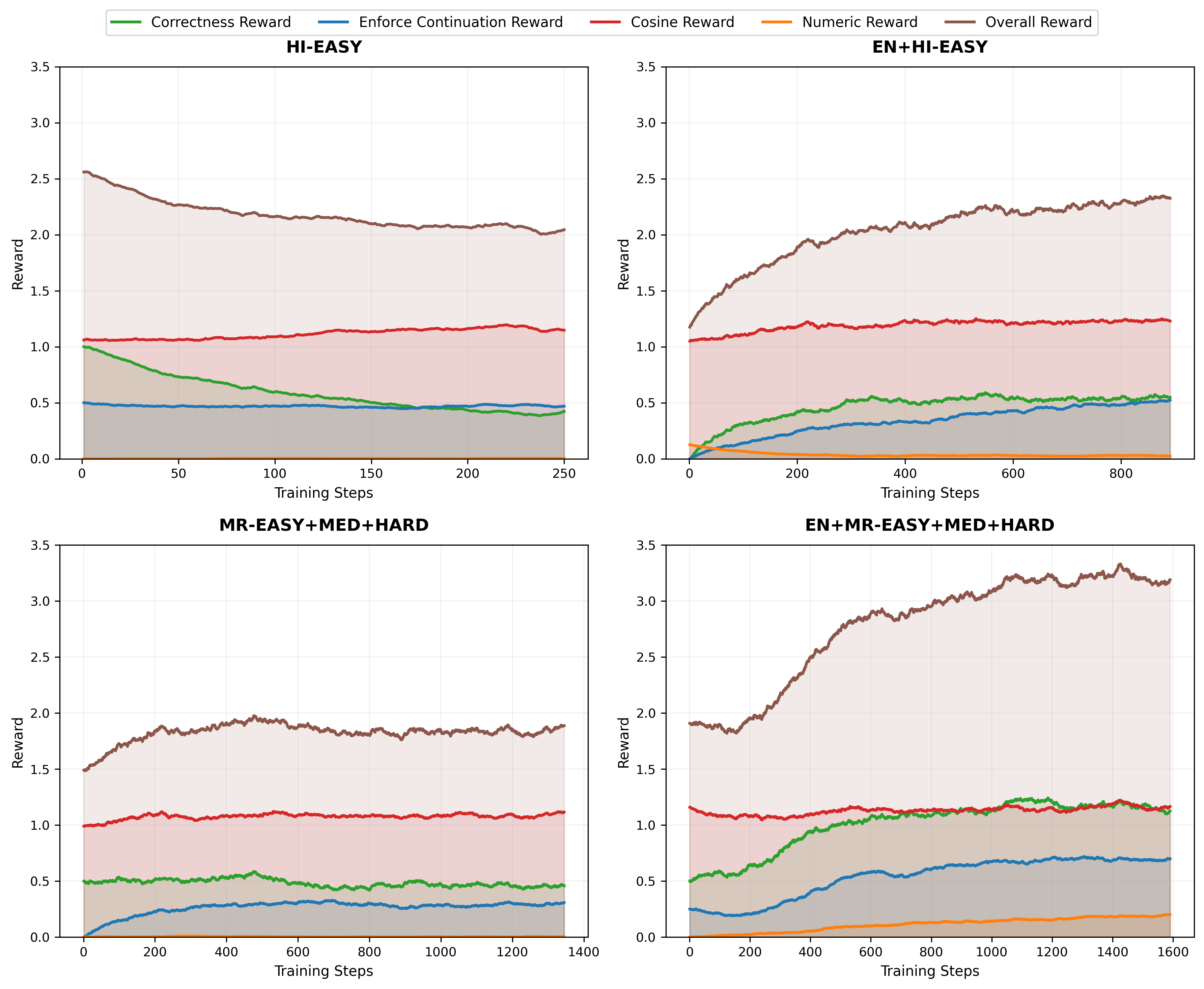}
    \caption{\textbf{Reward curves comparing monolingual (Hindi, Marathi) and bilingual (English+Hindi, English+Marathi) training.} Bilingual setups show faster convergence, higher overall reward, and more stable training, while monolingual models plateau early. This highlights the role of English as a regularizer, improving reasoning quality and stability in low-resource languages.}
\label{fig:reward_lang}
\end{figure}

Figure~\ref{fig:reward_lang} validates this directly and unambiguously. In the Hindi-only setting (HI-EASY), overall reward declines from approximately 2.5 to 2.0 across just 250 steps, consistent with a model struggling to simultaneously maintain reasoning quality while adapting to an unfamiliar script distribution. Adding English (EN+HI-EASY) reverses this entirely: reward rises steadily from 1.0 to 2.3 across 900 steps, with a qualitatively smoother trajectory throughout. The effect is even more pronounced for Marathi. Training on Marathi alone produces reward that stagnates between 1.5 and 2.0 across 1,400 steps with no meaningful upward trend, while EN+MR climbs consistently from approximately 1.8 to 3.2 across 1,600 steps, the highest overall ceiling observed across all configurations. English augmentation does not merely contribute additional training signal but it fundamentally changes the nature of the learning dynamic from stagnant or declining to monotonically increasing. This confirms that anchoring reasoning in English is not an auxiliary benefit but a necessary condition for stable and effective multilingual RL post-training. 

\section{Ablation Study} 
\label{sec:ablation}

\subsection{Mix-Ratio Sweep}
We trained three English–Marathi variants by varying the English fraction (Table~\ref{tab:mixratio}). A 20\% English proportion consistently yields the best performance across all difficulty levels, suggesting that a small English component acts as a regularizer, broadening the learning signal while maintaining Marathi as the dominant language.
We therefore fix the English–Marathi ratio to 80–20 in subsequent experiments.
Using this fixed mixture, we also compare \textbf{IRIS} against vanilla GRPO baseline to isolate its effects from data mixing.

\begin{table}[htbp]
\centering
\small
\caption{ Performance of English–Marathi mix ratios on Qwen2.5-Math. The 80/20 mix achieves the best balance.}
\resizebox{\columnwidth}{!}{

\begin{tabular}{lccc}
\toprule
\textbf{English Mix \%} & \textbf{Easy} & \textbf{Easy+Med} & \textbf{Easy+Med+Hard} \\
\midrule
0\%  & 41.0 & 57.5 & 59.0 \\
20\%(\textbf{IRIS}) & \textbf{57.5} & \textbf{59.3} & \textbf{64.2} \\
20\%(\textbf{vanilla GRPO}) & 49.0 & 55.7 & 56.7\\
50\% & 50.6 & 50.6 & 59.0 \\

\bottomrule
\end{tabular}
}
\label{tab:mixratio}
\end{table}

\subsection{Curriculum and Reward Ablations}
Table~\ref{tab:qwen_ablation} evaluates the impact of curriculum structure and reward design on Qwen2.5-Math-7B. Skipping SFT warm-up (\textbf{H Only}) or eliminating curriculum progression (\textbf{V+H No Curriculum}) reduces accuracy, highlighting the importance of vertical initialization and staged difficulty exposure.

Simplifying the reward to correctness-only similarly degrades performance, with further drops when cosine alignment is removed (Easy accuracy: 87.0\% $\rightarrow$ 84.3\%). Direct training on hard-level data without curriculum staging achieves only 80.9\%, confirming that progressive scaffolding is essential. The full (V+H) configuration, combining staged curriculum with composite reward, performs best across all difficulty levels.

\begin{table}[htbp]
\centering
\small
\setlength{\tabcolsep}{3pt}
\caption{Impact of curriculum progression and reward composition on Qwen2.5-Math.}

\label{tab:qwen_ablation}
\resizebox{\columnwidth}{!}{
\begin{tabular}{lccc}
\toprule
\textbf{Setting} & \textbf{Easy} & \textbf{Easy+Med} & \textbf{Easy+Med+Hard} \\
\midrule
H Only & 81.8 & 79.5 & 79.5 \\
V (No Curriculum) & -- & -- & 85.1 \\
V+H (No Curriculum) & -- & -- & 86.5 \\
V+H (Correctness) & 85.0 & 86.0 & 81.2 \\
V+H (Full) & \textbf{87.0} & \textbf{88.4} & \textbf{85.3} \\
\bottomrule 
\end{tabular}
}
\end{table}

\subsection{PPO-Only Reverse Curriculum}
\label{sec:ppo-ablation}

We evaluate a PPO-only reverse curriculum baseline inspired by R$^3$~\citep{tao2024reverseforwardcurriculumlearning}, which omits stage-wise weighting, cosine alignment, and auxiliary rewards. This setup represents a raw combination of PPO and reverse curriculum without our proposed composite reward shaping. Despite comparable training duration, this baseline reaches 82.1\% on \emph{Easy} and 82.7\% on \emph{Easy+Medium}, but drops to 76.5\% on \emph{Easy+Medium+Hard}, indicating that curriculum scheduling alone is insufficient and that the structured RLVR composite reward is required for stable performance at higher difficulty.

\section{Overview of Results}
\label{sec:results}

We report results on the \emph{Medium} test split, which offers a balanced setting for evaluating mathematical reasoning without saturation on easy problems or instability on the hardest ones. For Hindi and Marathi, English test items are translated to ensure strict cross-lingual equivalence.

\paragraph{Results on CL-Math.}
Table~\ref{tab:final_results} reports accuracy across curriculum settings. Along the \emph{vertical axis}, expanding training from Easy → Easy+Medium → Easy+Medium+Hard consistently improves performance: \textbf{Qwen2.5-Math-7B} increases from 85.2 → 86.3 on English, 52.7 → 57.7 on Hindi, and 38.3 → 55.7 on Marathi, with \textbf{WizardMath-7B} showing the same trend. 

\begin{figure}
    \centering
    \includegraphics[width=\linewidth]{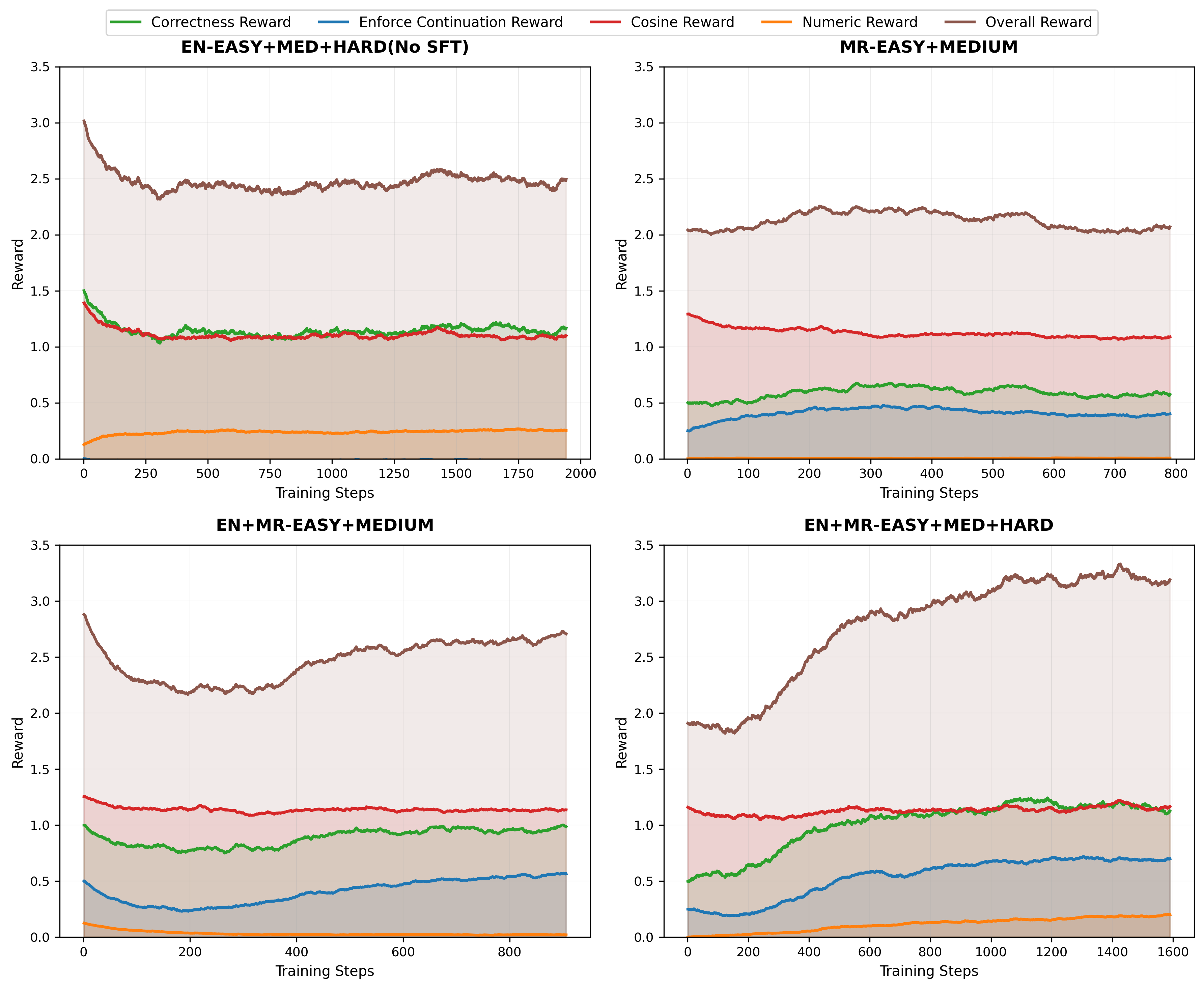}
    \caption{\textbf{Training reward dynamics across curriculum configurations.}
Without SFT warmup, rewards plateau early (${\sim}2.5$). Monolingual Marathi on Easy+Medium stabilizes at ${\sim}2.1$, while adding English improves performance to ${\sim}2.7$ but saturates at medium difficulty. The full IRIS setup (Easy+Medium+Hard) achieves sustained gains up to ${\sim}3.2$, showing that SFT warmup, bilingual mixing, and full curriculum each incrementally raise the performance ceiling.}
\label{fig:reward_progression}
\end{figure}

Adding the \emph{horizontal reinforcement stage} (V+H) further boosts accuracy, with modest gains on English (+1–3\%) and larger improvements on Hindi (+12–16) and Marathi (+3–4); the slight dip at Easy+Medium+Hard likely reflects the single-epoch training limit. In bilingual settings, English–Hindi and English–Marathi mixing amplifies performance, with Qwen’s Hindi score rising from 57.7 → 73.5 (+15.8\%) and WizardMath’s Marathi score improving from 18.6 → 20.6. Figure~\ref{fig:reward_progression} illustrates the contribution of each IRIS component through training reward dynamics.

\paragraph{Results on External Benchmarks.}

Table~\ref{tab:benchmark_results} extends evaluation to SVAMP, GSM8K, and MATH. On standard English benchmarks, Our curriculum-trained \textbf{Qwen2.5-Math-7B (IRIS)} achieves 90.6\% on SVAMP, 83.9\% on GSM8K, and 64.6\% on MATH, representing +8.5, +4.6, and +9.2 \% gains respectively over the base model. While GPT-4 remains strongest, IRIS substantially outperforms other specialized 7B models (MetaMath, DeepSeek-Math) across all three benchmarks.
\textbf{WizardMath-7B (IRIS)} mirrors this trajectory, consistently outperforming its base variant. These results confirm that the proposed vertical+horizontal curriculum generalizes beyond CL-Math, strengthening reasoning across languages, difficulty tiers, and benchmarks.

\section{Conclusion}
We introduced IRIS (Interleaved Reinforcement with Incremental Staged Curriculum), a two-axis training framework that combines staged supervised fine-tuning over increasing difficulty with reverse curriculum reinforcement learning. By separating skill acquisition and reasoning refinement along vertical and horizontal axes, IRIS enables stable learning of long-horizon reasoning. Across English, Hindi, and Marathi, the framework consistently outperforms single-axis baselines, with particularly strong gains in low-resource and bilingual settings. IRIS further generalizes to standard mathematical reasoning benchmarks, demonstrating that structured curriculum design offers a simple yet effective approach for multilingual reasoning.

Beyond these empirical gains, our results carry a broader implication for multilingual RL post-training. The consistent advantage of English augmentation as even 20\% of the training mix acts as a high-resource reasoning anchor. It is not merely beneficial but structurally necessary when the target language lacks sufficient data to independently sustain reward signal. This decoupling of reasoning acquisition from script 
adaptation may extend beyond Devanagari script. Future work will investigate extend IRIS to non-mathematical reasoning domains, 
such as multi-step logical inference or code generation, we would test whether the vertical-horizontal curriculum decomposition is 
domain-agnostic and reward functions scale beyond mathematical 
correctness.



\section{Limitations}

Our work has several limitations. First, computational constraints limit training to a single epoch per curriculum stage, which may hinder full convergence on harder tiers and contribute to the modest gap between Easy+Med and Easy+Med+Hard performance. Second, CL-Math includes only 29k samples, restricting large-scale multilingual evaluation. Third, the curriculum follows a fixed Easy→Medium→Hard progression, which may not be optimal for models with different initial capabilities or learning dynamics. Finally, experiments are limited to 7B-parameter models, and scalability to larger models remains an open question.

\bibliography{custom}

\section{Appendix}
\label{sec:appendix}

This Technical Appendix is a supplement to “IRIS: Interleaved Reinforcement with Incremental Staged Curriculum for Cross-Lingual Mathematical Reasoning”. The following sections include a detailed glossary of our Methodology, Algorithm, Prompts used at different stages of training, Evaluation Explanations, Computational Resources and Software Environment, Train Logs Analysis continued from the main paper, Qualitative Evidence of Curriculum Progression and Translation quality from English to Indian languages.

\section{Methodology Glossary}
\begin{description}
  \item[CL-Math] Our newly-curated multilingual maths corpus. It extends \textit{IndiMathQA} by adding (i) step-by-step solutions and (ii) Hindi \& Marathi translations, so models can practise both reasoning depth and cross-lingual transfer.
  \item[Curriculum Learning (CL)] A training strategy that presents easier examples first and progressively introduces harder ones, creating a smooth learning trajectory. In our vertical axis this corresponds to Easy $\rightarrow$ Medium $\rightarrow$ Hard.
  \item[Step-by-Step (Chain-of-Thought) Solutions] Full, numbered reasoning traces for each problem; they expose the intermediate logical steps rather than only the final answer.
  \item[LoRA Adapter] A lightweight low-rank adaptation layer injected into a pretrained model, enabling efficient fine-tuning by updating a small fraction of parameters.
  \item[Llama 3.3 (70B)] The 70-billion-parameter base model used for both dataset augmentation and as the external evaluator in the automatic grading stage.
  \item[Supervised Fine-Tuning (SFT)] A standard teacher-forcing approach in which the model is trained to predict the next token in reference solution traces using cross-entropy loss. We apply SFT incrementally, starting with Easy problems, then Easy+Medium, and finally Easy+Medium+Hard settings, allowing the model to gradually learn reasoning patterns of increasing complexity.

  \item[Step-Wise Continuation (Horizontal Axis)] A reverse/continuation curriculum where the model receives only a shrinking prefix of the solution and must generate the remainder, forcing deeper planning from reduced context.
  \item[GRPO (Group Relative Policy Optimisation)] The reinforcement-learning algorithm used on the horizontal axis. It evaluates samples relative to each other from the same prompt, yielding more stable gradients than independent scoring.
  \item[Cosine-Alignment Reward] A component of the RL reward measuring cosine similarity between generated continuations and reference suffixes; it provides partial credit for semantically aligned but lexically varied reasoning.
  \item[IndicTrans 2] The AI4Bharat translation model used to convert English problems and structured solutions into Hindi and Marathi while preserving step-level coherence.
  \item[Bilingual Cross-Transfer] Training on a balanced mixture (50\% English, 50\% Indian language) so that knowledge learned in one script can transfer and reinforce performance in the other.
  \item[RLVR] A framework that decomposes the reward signal into transparent, verifiable components, such as answer correctness, reasoning alignment, and format consistency enabling stable and interpretable policy improvement.
\end{description}

\subsection{Prompts}
\subsubsection{CL-Math dataset augmentation}
\begin{figure}[htbp]
  \centering
  \includegraphics[width=\linewidth]{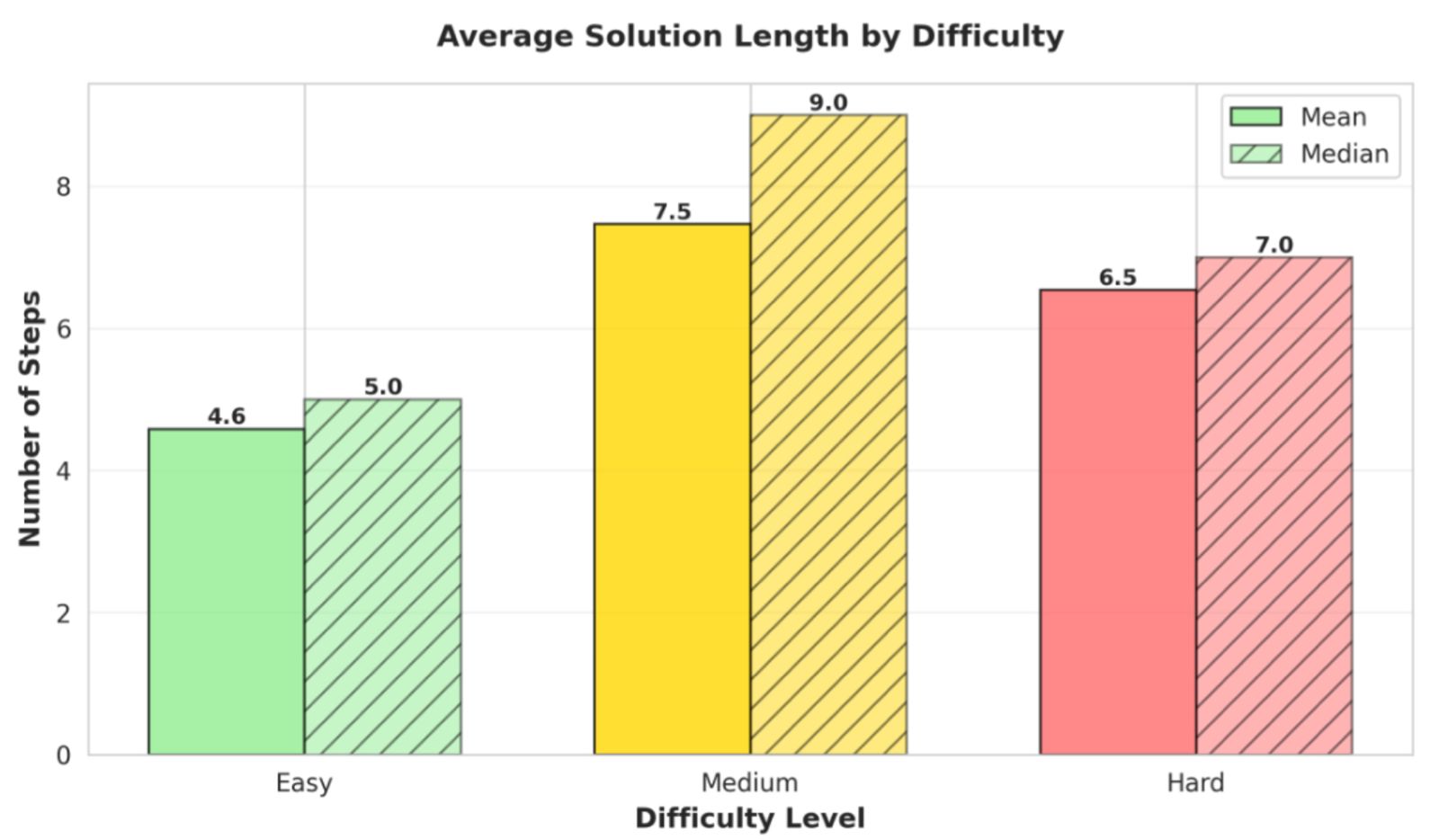} 
  \caption{\textbf{Average Solution Length by Difficulty:} Comparison of mean and median number of reasoning steps generated per difficulty level in the CL-Math dataset.}
  
  \label{fig:avg_solution_length}
\end{figure}

We use the Chain-Of-Thought prompting technique to generate reasoning steps for the entire dataset with variations depending on the difficulty. The input contains a question and its solution, which is converted into a specific number of logical steps with consistent transitions between them. Figure~\ref{fig:avg_solution_length} shows the average solution length distribution across difficulty levels. The number of reasoning steps increases progressively from Easy to Hard problems, confirming that the prompt design effectively controls the reasoning depth during dataset augmentation.

\begin{enumerate}

\item \textbf{Easy}
\begin{quote}
\small\texttt{
You have the following question and its corresponding answer.\\
Your task is to convert the answer only into 3 - 5 logical steps.\\
\# Question: \{question\}\\
\# Answer: \{answer\}\\
Give the response in the following format:\\
\# Step wise format: [your response]
}
\end{quote}

\item \textbf{Medium}
\begin{quote}
\small\texttt{
You have the following question and its corresponding answer.\\
Your task is to convert the answer only into 5 - 7 logical steps.\\
\# Question: \{question\}\\
\# Answer: \{answer\}\\
Give the response in the following format:\\
\# Step wise format: [your response]
}
\end{quote}

\item \textbf{Hard}
\begin{quote}
\small\texttt{
You have the following question and its corresponding answer.\\
Your task is to convert the answer only into 7-9 logical steps.\\
\# Question: \{question\}\\
\# Answer: \{answer\}\\
Give the response in the following format:\\
\# Step wise format: [your response]
}
\end{quote}
\end{enumerate}
\subsubsection{Horizontal Axis : Step-Wise \emph{Continuation} Curriculum}
To support the horizontal curriculum’s objective of progressively training the model to complete longer segments of step-wise reasoning, a fixed system prompt is used throughout training and inference. This prompt clearly marks the point at which the model must begin generating its own reasoning, following a set of initial steps provided as context.

\textbf{System Prompt}
\begin{quote}
\emph{You are a maths question solving model, currently you are learning to be better. Following the instruction carefully:}

\emph{If the user message contains the token \texttt{<CONTINUE>}, that token marks the point where your reasoning must start. Continue from there, then answer.}
\end{quote}


\section{Evaluation Explanations}
We used \textbf{Llama 3.3-70B} as an automatic judge, comparing model outputs to gold solutions and returning true/false verdicts, which we aggregated for accuracy and error analysis.

\subsection{Evaluation Prompt}
\begin{quote}
\emph{Compare these two mathematical solutions and determine if they have the same final numerical answer.}\newline
\emph{First, identify the final numerical answer from each solution, then state if they are the same.} \newline
\emph{First solution: \texttt{predicted answer}}
\emph{Second solution: \texttt{true answer}}\newline
\emph{Please respond in this format:}\newline
\emph{First answer: [state the final numerical answer from first solution]}
\emph{Second answer: [state the final numerical answer from second solution]}
\emph{Are they equal: [true/false]}\newline
\emph{Reason: [briefly explain why they are equal or different]}

\end{quote}

\subsection{Human Agreement Analysis}
Dataset validation and evaluation reliability are assessed separately. For dataset construction, we compute Fleiss Kappa = 0.71 across four annotators, indicating substantial agreement (Section 3.1). For evaluation, we validate the Llama-70B judge against human annotations on 300 randomly sampled model outputs, obtaining Cohen’s Kappa = 0.795, confirming strong alignment between automatic and human judgments.

\subsection{Formatting and Parsing Consistency}
CL-Math uses a uniform step format across English, Hindi, and Marathi: reasoning steps are indexed using standard Arabic numerals (0–9) and separated by newline characters. As a result, integer-continuation and step-continuation rewards use identical parsing logic across languages. 

Language-specific prompt suffixes are handled via lightweight regex extraction prior to reward computation, ensuring consistent scoring without reliance on script-specific numeral systems.

\subsection{Dataset Validation}
CL-Math was validated by four annotators: two undergraduate and two master’s students, all with technical backgrounds. Given that the dataset targets high-school–level mathematics, annotators were sufficiently qualified to verify step-level correctness and solution consistency. Inter-annotator agreement is measured using Fleiss’ Kappa = 0.71, indicating substantial agreement.

\section{Computational Resources and Software Environment}
All experiments were run on a high-performance setup with hardware and software tailored for large language model training and RL. Tables~\ref{tab:hardware} and~\ref{tab:software} summarize the components used.

\begin{table}[h]
\centering
\caption{\textbf{Hardware Specifications.}}
\resizebox{\columnwidth}{!}{
\begin{tabular}{ll}
\hline
\textbf{Component} & \textbf{Specification} \\
\hline
GPU & NVIDIA H200 (144 GB), A100 (40 GB) \\
CPU & AMD EPYC 7742, 64 cores \\
RAM & 512 GB \\
Operating System & Ubuntu 22.04 LTS \\
\hline
\end{tabular}
}
\label{tab:hardware}
\end{table}

\begin{table}[h]
\centering
\caption{\textbf{Software Environment.}}
\resizebox{\columnwidth}{!}{
\begin{tabular}{ll}
\hline
\textbf{Library} & \textbf{Version} \\
\hline
Python & 3.10 \\
PyTorch & 2.7.0 \\
Transformers (HuggingFace) & 4.53.0 \\
TRL & 0.19.0 \\
PEFT & 0.12.0 \\
Unsloth & Git (latest) \\
VLLM & Compatible release \\
Datasets & 3.6.0 \\
SentenceTransformers & 2.6.1 \\
\hline
\end{tabular}
}
\label{tab:software}
\end{table}

\subsection{Cross-Lingual Transfer Effects}

Introducing a small amount of English into the Marathi pipeline makes the learning trajectory both stronger and more durable. The mixed model doesn’t stall early like the monolingual run; instead it sustains useful gradient signal longer, yielding smoother reward growth. Intermediate signals, especially correctness, improve while noise does not increase, suggesting the additional language enriches the model’s representation space without destabilizing training. In practice, English acts as a complementary regularizer, nudging the policy away from language-specific quirks and into more robust reasoning behavior.

\subsection{Curriculum-Driven Reward Enhancement}

Advancing from medium to hard problems with the staged curriculum yields clear qualitative gains. The model not only achieves higher average returns but also suppresses low-performing trajectories and continues improving late in training, unlike the medium-only variant which plateaus. Final answer correctness strengthens in tandem, indicating that the curriculum’s gradual escalation both deepens and stabilizes the model’s reasoning capabilities.

By design, the cosine reward stays flat: its linearly decaying stage weight exactly offsets any similarity gains.

\begin{figure*}[htbp]
    \centering
    \includegraphics[width=1\linewidth]{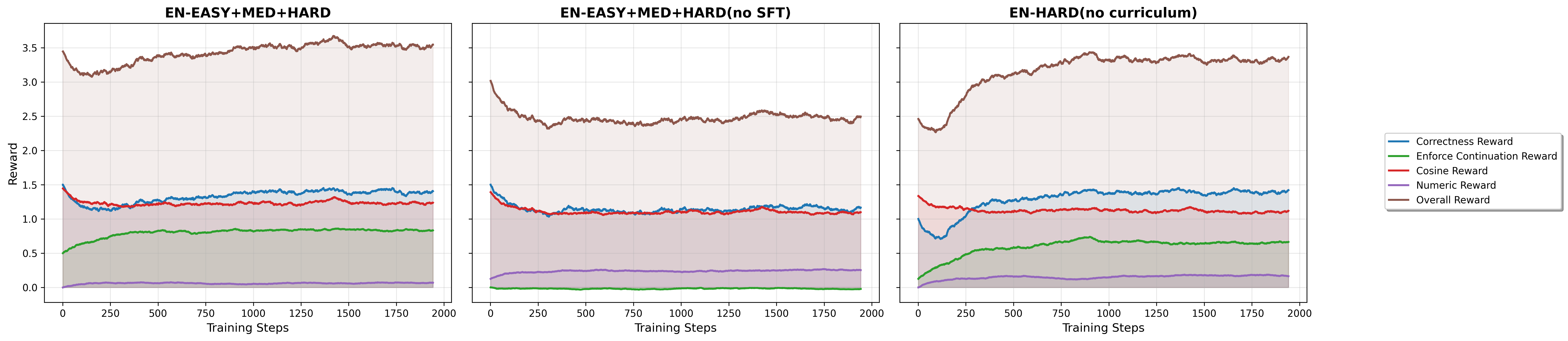}
    \caption{We compare reward trends across three configurations: full curriculum with SFT (left), curriculum without SFT initialization (center), and training only on hard examples without any curriculum (right). The full curriculum consistently achieves higher overall and correctness rewards, highlighting the importance of both gradual progression and supervised initialization in stabilizing and improving training.}
    \label{fig:eng}
\end{figure*}

\begin{figure*}[htbp]
    \centering
    \includegraphics[width=1\linewidth]{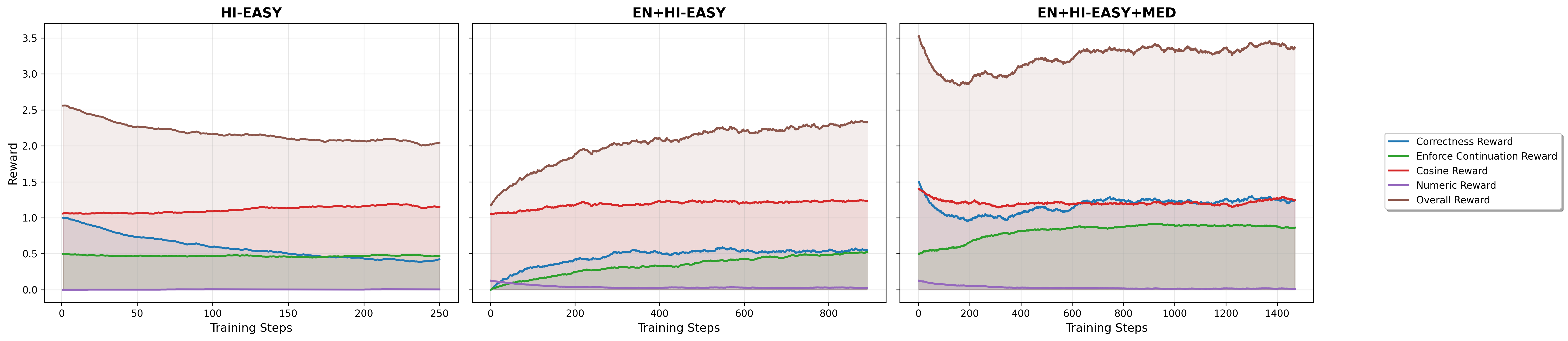}
    \caption{This figure illustrates reward progression when training on Hindi-only data (left), English+Hindi (center), and English+Hindi with curriculum (right). The cross-lingual setups significantly outperform the monolingual Hindi baseline, especially when combined with curriculum, demonstrating strong transfer from English supervision and the value of structured progression in multilingual alignment.}
    \label{fig:hind}
\end{figure*}

\begin{figure*}[htbp]
    \centering
    \includegraphics[width=1\linewidth]{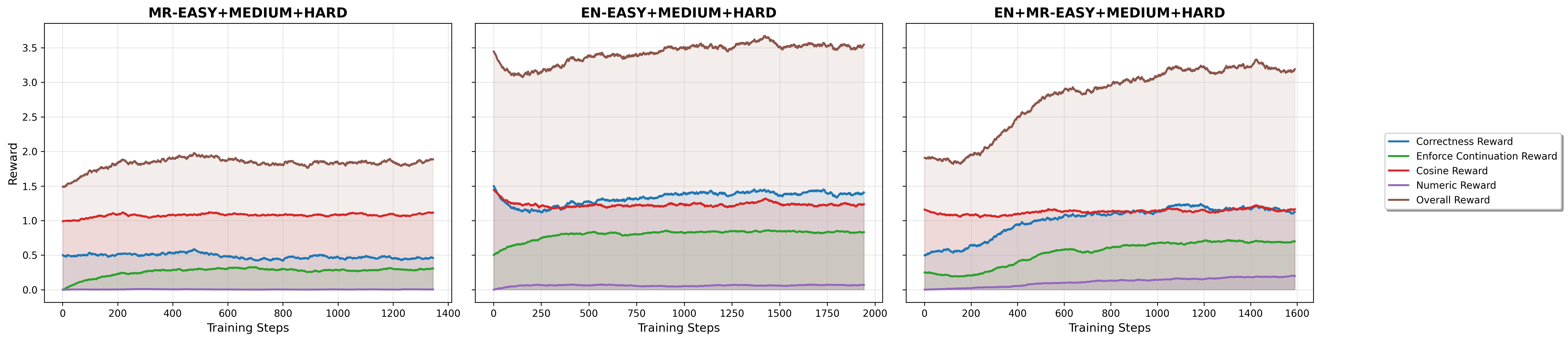}
    \caption{Reward curves for Marathi-only (left), English-only (center), and combined English+Marathi curriculum training (right) show that multilingual curriculum learning leads to superior reward optimization. The joint setting achieves the highest correctness and overall rewards compared to monolingual Marathi, underscoring the benefit of leveraging English data to support low-resource language training.}
    \label{fig:hard}
\end{figure*}

For completeness, we also include additional reward progression comparisons for English curriculum variations (see Figure \ref{fig:eng}), Hindi–English cross-lingual (see Figure \ref{fig:hind}) training settings, and Hard Tier Models in English and Marathi (see Figure \ref{fig:hard}), analogous to the Marathi–English Curriculum analyses presented in the main paper.

\section{Qualitative Evidence of Curriculum Progression}

To demonstrate the value of successive curriculum stages, we show representative final answers for the same mathematical problems under different checkpoints. Rows where only the earlier tier fails but the next tier succeeds (e.g., Easy \(\rightarrow\) Medium) are intended to highlight that the intermediate curriculum slice contributes nontrivial new capability rather than being redundant(see Figures \ref{fig:combined_analysis1} through \ref{fig:combined_analysis14}).

\section{Translation Quality}

We additionally include a translation quality section, presenting side-by-side comparisons of the original English prompts with their Marathi and Hindi translations to illustrate clarity across languages (see Figures \ref{fig:combined_analysis15}, \ref{fig:combined_analysis16} and \ref{fig:combined_analysis17}).

\section{GRPO Loss Formulation}
\label{appendix:grpo}

For each problem $(q, \mathbf{y})$ at stage $k$, we sample $G$ rollouts:
\begin{equation}
\{\hat{\mathbf{y}}_k^{(1)}, \ldots, \hat{\mathbf{y}}_k^{(G)}\} \sim \pi_\theta(\cdot \mid \mathbf{x}_k)
\end{equation}
where $\mathbf{x}_k = [q, s_1, \ldots, s_{m_k}, \texttt{<CONTINUE>}]$.

Each rollout's advantage is computed relative to the group:
\begin{equation}
A_k^{(i)} = R_k(q, \hat{\mathbf{y}}_k^{(i)}) - \frac{1}{G}\sum_{j=1}^{G} R_k(q, \hat{\mathbf{y}}_k^{(j)})
\end{equation}

The GRPO loss for stage $k$ is:
\[
\mathcal{L}_{\text{GRPO}}^{(k)}(\theta) = -\mathbb{E}_{(q,\mathbf{y}) \sim \mathcal{D}}\left[ \frac{1}{G} \sum_{i=1}^{G} A_k^{(i)}\log \pi_\theta(\hat{\mathbf{y}}_k^{(i)} \mid \mathbf{x}_k) \right] \]\[+ \text{KL penalty}\]

where the KL penalty term prevents excessive deviation from the reference policy $\pi_{\text{ref}}$. We use $\beta = 0.01$ as the KL coefficient.

This follows standard GRPO \citep{shao2024deepseekmath}. Our contribution is the curriculum-weighted combination across stages.

\begin{figure*}
  \centering
  
  \begin{minipage}{\textwidth}
    \centering
    \includegraphics[width=\linewidth]{images/1.jpg}
  \end{minipage}

  \begin{minipage}{\textwidth}
    \centering
    \includegraphics[width=\linewidth]{images/2_1.jpg}
  \end{minipage}

  \begin{minipage}{\textwidth}
    \centering
    \includegraphics[width=\linewidth]{images/3.jpg}
  \end{minipage}
  
  \caption{Question and Base Model Response Analysis}
  \label{fig:combined_analysis1}
\end{figure*}

\begin{figure*}
  \centering
  
  \begin{minipage}{\textwidth}
    \centering
    \includegraphics[width=\linewidth]{images/4.jpg}
  \end{minipage}
  \label{fig:combined_analysis2}
\end{figure*}

\begin{figure*}
  \centering
  
  \begin{minipage}{\textwidth}
    \centering
    \includegraphics[width=\linewidth]{images/5.jpg}
  \end{minipage}
  
  \caption{\textbf{Comparison of Arithmetic Product Tasks on Vertical Curriculum:} We compared responses by Base Model and SFT responses of all difficulty tiers}
  \label{fig:combined_analysis3}
\end{figure*}

\begin{figure*}
  \centering
  
  \begin{minipage}{\textwidth}
    \centering
    \includegraphics[width=\linewidth]{images/6.jpg}
  \end{minipage}

  \begin{minipage}{\textwidth}
    \centering
    \includegraphics[width=\linewidth]{images/7.jpg}
  \end{minipage}
  \label{fig:combined_analysis4}
\end{figure*}

\begin{figure*}
  \centering

  \begin{minipage}{\textwidth}
    \centering
    \includegraphics[width=\linewidth]{images/8.jpg}
  \end{minipage}
  
 \caption{\textbf{Comparison of Quadratic Factorisation Task on Vertical Curriculum:} We compared responses by Base Model and SFT responses of easy and medium tiers}
  \label{fig:combined_analysis5}
\end{figure*}

\begin{figure*}
  \centering

  \begin{minipage}{\textwidth}
    \centering
    \includegraphics[width=\linewidth]{images/9.jpg}
  \end{minipage}
  \begin{minipage}{\textwidth}
    \centering
    \includegraphics[width=\linewidth]{images/10.jpg}
  \end{minipage}
  \label{fig:combined_analysis6}
\end{figure*}
\begin{figure*}
  \centering
  
  \begin{minipage}{\textwidth}
    \centering
    \includegraphics[width=\linewidth]{images/11.jpg}
  \end{minipage}
  
  \caption{\textbf{Comparison of Arithmetic Product Task on Horizontal Curriculum:} We compared responses after applying GRPO on respective SFT checkpoints.}
  \label{fig:combined_analysis7}
\end{figure*}

\begin{figure*}
  \centering
  
  \begin{minipage}{\textwidth}
    \centering
    \includegraphics[width=\linewidth]{images/12.jpg}
  \end{minipage}

  \begin{minipage}{\textwidth}
    \centering
    \includegraphics[width=\linewidth]{images/13.jpg}
  \end{minipage}
  \label{fig:combined_analysis8}
\end{figure*}

\begin{figure*}
  \centering
  
  \begin{minipage}{\textwidth}
    \centering
    \includegraphics[width=\linewidth]{images/14.jpg}
  \end{minipage}
  \label{fig:combined_analysis9}
\end{figure*}
\begin{figure*}
  \centering
  
  \begin{minipage}{\textwidth}
    \centering
    \includegraphics[width=\linewidth]{images/15.jpg}
  \end{minipage}
  
  \caption{\textbf{Comparison of Quadratic Factorisation Task on Horizontal Curriculum:} We compared responses after applying GRPO on respective SFT checkpoints.}
  \label{fig:combined_analysis10}
\end{figure*}

\begin{figure*}
  \centering
  
  \begin{minipage}{\textwidth}
    \centering
    \includegraphics[width=\linewidth]{images/22.jpg}
  \end{minipage}
  \begin{minipage}{\textwidth}
    \centering
    \includegraphics[width=\linewidth]{images/23.jpg}
  \end{minipage}
  \begin{minipage}{\textwidth}
    \centering
    \includegraphics[width=\linewidth]{images/24.jpg}
  \end{minipage}
  \label{fig:combined_analysis11}
\end{figure*}
\begin{figure*}
  \centering
  
  \begin{minipage}{\textwidth}
    \centering
    \includegraphics[width=\linewidth]{images/25.jpg}
  \end{minipage}
  \label{fig:combined_analysis12}
  \caption{\textbf{Comparison of Hindi Responses for Arithmetic Average Task on Horizontal Curriculum:} We compared responses after applying GRPO on respective SFT checkpoints.}
\end{figure*}

\begin{figure*}
  \centering
  
  \begin{minipage}{\textwidth}
    \centering
    \includegraphics[width=\linewidth]{images/26.jpg}
  \end{minipage}
  \begin{minipage}{\textwidth}
    \centering
    \includegraphics[width=\linewidth]{images/27.jpg}
  \end{minipage}
  \begin{minipage}{\textwidth}
    \centering
    \includegraphics[width=\linewidth]{images/28.jpg}
  \end{minipage}
  \label{fig:combined_analysis13}
\end{figure*}

\begin{figure*}
  \centering
  
  \begin{minipage}{\textwidth}
    \centering
    \includegraphics[width=\linewidth]{images/29.jpg}
  \end{minipage}
  
  \caption{\textbf{Comparison of Marathi Responses for Algebra Task on Horizontal Curriculum:} We compared responses after applying GRPO on respective SFT checkpoints.}
  \label{fig:combined_analysis14}
\end{figure*}

\begin{figure*}
  \centering
  
  \begin{minipage}{\textwidth}
    \centering
    \includegraphics[width=\linewidth]{images/16.jpg}
  \end{minipage}
  \begin{minipage}{\textwidth}
    \centering
    \includegraphics[width=\linewidth]{images/17.jpg}
  \end{minipage}
  
  \caption{Original English Arithmetic Question and Corresponding Generated Stepwise Response}
  \label{fig:combined_analysis15}
\end{figure*}
\begin{figure*}
  \centering
  
  \begin{minipage}{\textwidth}
    \centering
    \includegraphics[width=\linewidth]{images/18.jpg}
  \end{minipage}
  \begin{minipage}{\textwidth}
    \centering
    \includegraphics[width=\linewidth]{images/19.jpg}
  \end{minipage}
  
  \caption{Marathi Translation of the English Arithmetic Question and the Corresponding Generated Stepwise Response}
  \label{fig:combined_analysis16}
\end{figure*}

\begin{figure*}
  \centering
  
  \begin{minipage}{\textwidth}
    \centering
    \includegraphics[width=\linewidth]{images/20.jpg}
  \end{minipage}
  \begin{minipage}{\textwidth}
    \centering
    \includegraphics[width=\linewidth]{images/21.jpg}
  \end{minipage}
  
  \caption{Hindi Translation of the English Arithmetic Question and the Corresponding Generated Stepwise Response}
  \label{fig:combined_analysis17}
\end{figure*}

\end{document}